# Adversarial Attacks on Machine Learning in Embedded and IoT Platforms

Christian Westbrook, Sudeep Pasricha

*Abstract*—Machine learning (ML) algorithms are increasingly being integrated into embedded and IoT systems that surround us, and they are vulnerable to adversarial attacks. The deployment of these ML algorithms on resource-limited embedded platforms also requires the use of model compression techniques. The impact of such model compression techniques on adversarial robustness in ML is an important and emerging area of research. This article provides an overview of the landscape of adversarial attacks and ML model compression techniques relevant to embedded systems. We then describe efforts that seek to understand the relationship between adversarial attacks and ML model compression before discussing open problems in this area.

## I. Introduction

Deep neural networks (DNNs) are increasingly becoming a part of the embedded and IoT systems that surround us, from facial recognition in smartphones and voice recognition in smart home speakers to autonomous navigation and control in emerging vehicles. As the overlap between the domains of embedded systems and machine learning (ML) continues to increase, developers must find ways to deploy DNNs efficiently on resource-constrained embedded platforms. Unfortunately, state-of-the-art DNNs continue to push the boundaries of performance, which has led to a steady increase in model size and complexity. For example, natural language processing models such as GPT-3 [1] and computer vision models such as Vision Transformers (ViT) [2] possess billions of parameters. This trend presents challenges to applications of ML in embedded environments where resources such as memory and energy are tightly constrained. One active area of research aimed at reducing the memory footprint, energy expenditure, and inference latency of ML models for use in resource-constrained environments is the field of model compression.

Model compression techniques provide several key benefits to DNNs in embedded environments [3]. Compressed DNN models can have fewer and more compact parameters, and as such, they take up less space in memory and require fewer operations at inference time. The reduced computational intensiveness due to model compression also reduces latency and energy consumption. For example, the use of quantization and pruning-based model compression techniques allowed a DNN-based framework for indoor localization on embedded mobile devices to reduce model size by 81% and inference latency by 31%, at a minimal cost of 5% model accuracy reduction [4]. These improvements can enable the deployment of new ML applications that employ progressively more powerful models in embedded environments.

The growing number of ML models making their way into commodity embedded systems raises questions about their robustness to adversarial attacks. It was demonstrated in [5] that ML systems are vulnerable to inference-time evasion attacks involving small, imperceptible perturbations introduced into input samples that caused DNNs to misclassify their samples. Such forced misclassification could have severe consequences for embedded ML. Examples of the exploitation made possible by such attacks include an attacker causing a self-driving vehicle to misclassify a stop sign [6] and a seemingly innocuous object being misclassified as a firearm [7]. Adversarial attacks can also occur during training time, e.g., poisoning [8], [9] and backdoor attacks [10]. These discoveries highlight a need for research on the impact of adversarial attacks in a world where ML is becoming tightly intertwined with widely used embedded and IoT systems.

What happens when adversarial attacks target ML systems optimized for embedded environments? In [11], it was observed that because adversarial attacks often exploit the precision of the data they attack, a reduction in this precision could potentially render an attack less effective. Although model compression techniques modify the precision on which many adversarial attacks depend, recent research indicates that the relationships between model compression techniques and robustness to adversarial attacks are complex. A deeper understanding of these relationships is needed for a better assessment of the risks involved in deploying compressed ML models to embedded environments.

Adversarial attacks and model compression have been separately studied in prior work. The authors of [12] survey adversarial attacks and defenses but do not consider the impact of model compression on the attack scenarios. The authors of [3] survey strategies for ML model compression without considering the implications of compression on adversarial attacks. This article discusses techniques from the fields of adversarial machine learning and model compression before discussing recent research exploring the relationship between these areas. We also discuss open challenges and future directions relevant to emerging embedded and IoT systems.

## II. Adversarial Machine Learning

Adversarial machine learning is the study of attacks that attempt to degrade an ML system's ability to make accurate predictions. Although the phrase 'adversarial attack' is sometimes used exclusively to refer to evasion attacks, in this article, we broadly define adversarial attacks to be inclusive of evasion attacks, poisoning attacks, backdoor attacks, and exploratory attacks [9], [12].

*Evasion attacks* target the ML inference phase, seeking to evade the system by forcing the victim model to misclassify its input samples. The attack is accomplished by introducing malicious adjustments to input samples, known as adversarial perturbations. Various algorithms have been devised for generating the perturbation required to force misclassification.

Evasion attack approaches typically begin by estimating direction sensitivity to class change with respect to each dimension of the input sample considered as an input vector. These estimations are then used to inform the selection of a perturbation. The Fast Gradient Sign Method (FGSM) is an early example of this method [13]. The Basic Iterative Method (BIM) improves on FGSM by continuing to refine the perturbation iteratively [14]. But these methods do not consider the relative impact of perturbing a particular dimension, choosing instead to perturb every dimension. In contrast, the Jacobian Saliency Map Approach (JSMA) computes an adversarial saliency score for each dimension of the input vector and selects a pair of dimensions (one for each direction) to perturb that will have the largest predicted adversarial impact [15]. This process is repeated until the target sample is misclassified by the victim model, producing samples that sit just beyond the classification boundaries of the victim model. But the misclassification is very sensitive to precision changes and defense measures taken by the target system [11]. The Carlini and Wagner (C&W) attack addresses this problem with an improved objective function that also maximizes the difference between the victim model's confidence in the adversarial target class and the next most likely class [16]. All of these attacks seek to exploit knowledge of the model gradients and parameters to minimize the required adversarial perturbation and, as such, are considered white-box attacks. In contrast, in black-box attacks, such as the HopSkipJump and Square attacks [12], an adversary does not have access to the model structure, gradients, or parameters. White-box attacks are generally more effective than black-box attacks as the knowledge of internal model gradients and parameters makes it faster and more efficient to calculate the minimum change needed to create adversarial perturbations.

*Poisoning attacks* are untargeted attacks in which the attacker seeks to degrade model performance by undermining the training process. Data poisoning attacks introduce adversarial samples or perturb existing samples in the training dataset, causing undesirable changes in the decision boundaries learned by the model [8]. Alternatively, in model poisoning attacks, the model is poisoned directly during training by manipulating the cost function or the gradients computed during training to compromise model performance [9].

*Backdoor attacks* can be considered as a form of targeted poisoning attacks. Rather than seeking to degrade model performance in general, as done in poisoning attacks, here the attacker attempts to create a backdoor that allows adversarial input samples to trigger misclassification into a target class. An example in [10] demonstrates the insertion of a backdoor into a facial recognition system, despite the adversary lacking knowledge of either the victim model or the training set.

*Exploratory attacks* are unique in not attempting to degrade model performance. These attacks instead seek to learn as much as they can about the victim model by probing it with legitimate samples. In a model inversion exploratory attack, the probed information is used to infer sensitive details about the features present in the victim model's training dataset. One example is the reconstruction of the faces used to train a facial recognition system [17]. Model extraction exploratory attacks seek to use probing to construct a replica of the victim model. The replica can then be used to stage further attacks. In [18], the authors demonstrate that they can accomplish the same black-box model inversion attack performed in [17] with fewer queries by extracting the victim model and then performing a white-box inversion attack on the replica.

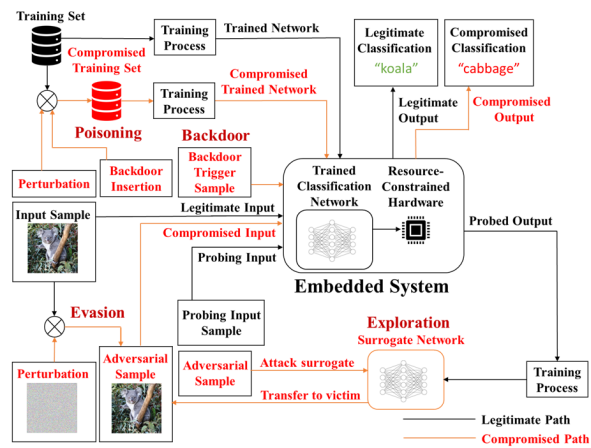

Fig. 1 Evasion, poisoning, backdoor, and exploratory attack scenarios.

In summary, evasion, poisoning, backdoor, and exploratory attacks (illustrated in Fig. 1) are major challenges for ML. These adversarial attacks have also been successfully transferred between models designed to perform similar tasks, even if their architectures or training sets are different [19]. This property dramatically complicates the challenge of defending ML models deployed in embedded systems against adversarial attacks, enabling scenarios in which exploratory attacks are used to extract the victim model and generate a surrogate model against which an evasion, poisoning, or backdoor attack can be crafted and then transferred back to the victim model.

## III. MODEL COMPRESSION

There are many potential applications of ML in embedded systems. Self-driving cars employ ML to detect lane markers, road signs, and pedestrians. Space systems use ML to curate and compress data for transmission to earth. Challenges common to these and other embedded environments usually involve resource constraints such as limited available memory, processing capability, or energy budget.

Model compression techniques can improve embedded ML performance in several ways. Compressed ML models require less memory, use less energy, and perform inference faster and with fewer computations. These benefits have led to the widespread adoption of model compression techniques in ML models deployed on embedded systems. Several model compression approaches have been particularly effective in reducing model resource requirements, including network pruning, quantization, and knowledge distillation [3].

*Network pruning* leverages a tendency towards sparsity in DNNs by removing weights, connections, filters, channels, or even entire layers that are unlikely to significantly impact model accuracy. Magnitude-based pruning approaches, for instance, consider parameters with values below a threshold as candidates for pruning. Various DNN pruning approaches have been demonstrated to produce significant reductions in model size with minimal impact on model accuracy [20].

*Network quantization* decreases ML model size by reducing the number of bits used to represent each parameter in the model [3]. One approach is to introduce quantization post-training and recover model performance with fine-tuning. Quantization-aware training is another approach that involves a simulation of model performance iteratively as weights are quantized gradually during training to minimize the accuracy lost by quantization [21]. Quantized models reduced to 8-bit integer or lower precision require far fewer and also simpler computations for inference than the floating-point operations necessary for their full-precision (32-bit) counterparts.

*Network knowledge distillation* uses principles of transfer learning to train a smaller model with the original model as a teacher [3]. One way to accomplish this is to use the logit vectors output by the original model as a target for the learner model to train against. In a more complex arrangement, a third model is interposed between the original and the learner, known as a 'teacher assistant' model, to facilitate a more effective transfer. The outcome of knowledge distillation is the generation of a smaller model that will consume less energy, take up less space in memory, and perform inference faster.

## IV. ATTACKING COMPRESSED MODELS

The impact of model compression techniques on adversarial robustness in ML is an important and emerging area of research. As model compression techniques become more widely adopted, it becomes essential to understand the relationship between each model compression technique and robustness to various forms of adversarial attacks.

In [22], the robustness of quantized DNNs against evasion attacks was explored. The authors performed an experiment assessing the robustness of CNNs trained for image classification and quantized to different levels of precision against several evasion attacks, including FGSM, BIM, and C&W attacks, among others. Based on their results, the authors argued that quantization leads to gradient masking, a state in which the model's gradients are rendered less useful to an attacker seeking to estimate direction sensitivity. But gradient masking is generally considered to be a weak defense, as, through model extraction, the original gradients can be reconstructed and attacked [23]. The robustness of binarized neural networks (BNNs) to evasion attacks was analyzed in [24]. In addition to the benefits of a reduced memory footprint and faster inference, the authors suggest that BNNs are more robust to evasion attacks than their full-precision counterparts in the best case and of equal robustness in the worst case.

In contrast, [25] suggests that quantized DNNs are even more vulnerable to evasion attacks due to a correlation between the bit width of the quantized model and an error amplification effect in which the classification error caused by the perturbation is amplified from layer to layer within the model. The authors argued that improvements to robustness due to quantization are nullified by the error amplification effect. They then proposed a way to regain the benefit by controlling the error amplification effect using a training technique called Defensive Quantization. This technique introduces Lipschitz regularization to the training process, effectively reducing the allowable rate of change in the value of the output of each layer, thereby controlling or reducing error amplification. This process was further improved in [26].

In [27], the authors proposed using network pruning to defend against evasion attacks which they call adversarial neural pruning. This process attempts to minimize the distortion introduced to the latent feature space by the evasion attack. Each latent feature is assessed for its vulnerability to distortion, and more vulnerable features are suppressed through the targeted pruning of network components.

In [28], the authors explored the impact of network pruning on robustness to poisoning attacks. They experimentally searched for a relationship between network pruning and robustness to poisoning attacks by assessing the robustness of a simple CNN pruned to different pruning ratios. The authors trained a CNN for image classification in each experiment and then iteratively pruned the trained model to a particular ratio. The model was then iteratively retrained with poisoned data for several epochs and tested for accuracy after each epoch. The authors found that although in every case the model eventually succumbs to degradation of performance, pruned models degrade at a much slower rate, requiring many additional epochs of poisoning than the original model.

The authors of [29] introduced a defense against evasion attacks using knowledge distillation and ensemble learning called Diverse Knowledge Distillation. The original model's knowledge is distilled into an ensemble of models with the caveat that each ensemble member is required to learn latent features distinct from the teacher. Inference is performed by polling each ensemble member and taking the majority vote. The authors called this idea Latent Space Separation. They argued that as the transferability of evasion attacks is reinforced by similarities in the learned latent spaces for vulnerable features, attacks on the teacher model will fail to transfer to the ensemble's unique assortment of latent feature spaces.

In [30], a technique for recovering from poisoning and backdoor attacks was proposed using knowledge distillation. The authors began by using an untrustworthy dataset to train a model that may have been poisoned or contain a backdoor. They then generated a new dataset of clean images without labels, called the distillation dataset, and attached predictions made by the untrusted model. A new distillation model was then trained using the distillation dataset. The untrusted and distillation models were both tasked with performing inference against the original untrustworthy dataset, and their predicted classes were compared. The authors argued that the input sample was clean when the distillation and untrusted models agreed with each other. When these models disagreed, the input sample might have been poisoned, as a poisoned model would generate an adversarial classification for poisoned samples. These potentially poisoned samples were removed from the dataset to create a detoxified dataset, and the distillation model was fine-tuned using this dataset. The resulting distilled model should have disabled any backdoors and recovered from damage due to poisoning attacks.

## V. OPEN CHALLENGES AND FUTURE DIRECTIONS

The studies discussed in the previous section explored the impact of adversarial attacks on compressed ML models and the use of model compression to defend against adversarial attacks. The relationships between adversarial attacks and model compression techniques must be further explored to better understand the security risks when deploying ML in embedded and IoT environments. This section discusses some potentially fruitful directions for future work in this field.

**Attacks on Model Compression Pipelines:** While various existing works assess how adversarial attacks perform against models compressed using a single compression technique, it would be desirable to consider the case when multiple model compression techniques are combined together in a model compression pipeline. Will attack scenarios based on chained model compression techniques (e.g., pruning and quantization [31]) retain the characteristics of attacks on each member technique? New work addressing the impact of different combinations of model compression techniques on robustness to adversarial attacks would be a beneficial contribution.

**Attacks on Complex Models and Datasets:** Many works presented in Section IV report experimental results based on attack scenarios involving relatively small DNN architectures or datasets. The attack scenarios presented in [22], [24], [28], [29], for instance, employ architectures with fewer than 15,000 neurons while architectures attacked in [25], [27], [30] contain millions. The attack scenarios presented in [22], [24], [25], [27]-[30] target models trained on small datasets such as CIFAR-10 and MNIST. Future work should experiment with state-of-the-art models trained on more complex datasets.

**Backdoor Attacks:** While there are examples of existing work exploring the impact that model compression techniques have on robustness to backdoor attacks [30], existing work has emphasized knowledge distillation [30] and network pruning [32]. Future work could further explore the relationship between quantization and robustness to backdoor attacks.

**Exploratory Attacks:** It would be beneficial to assess whether model compression impacts robustness to exploratory attacks. Would it take significantly more or fewer queries for an attacker to perform an extraction attack or an inversion attack against a compressed model? A change in the number of queries required to execute an exploratory attack would also impact the ability of a security system to detect that the attack is occurring.

**Attacks on Diverse Model Compression Approaches:** Existing work has primarily focused on analyzing the specific case of evasion attacks against quantized models. A gap seemingly exists between the volume of work being performed to improve our understanding of this scenario than other combinations of attacks and compression techniques. Future work could seek to close this gap by emphasizing neglected attack scenarios. In this article, we have emphasized network quantization, network pruning, and knowledge distillation as promising model compression techniques and discussed their relationships to robustness against adversarial attacks. Future work should consider additional compression techniques, including matrix approximation techniques (e.g., Tucker decomposition, CP decomposition) and the use of structured matrices (e.g., Circulant, Hadamard).

**Compression Defenses:** Although a wealth of research exists on defenses to adversarial attacks in ML, there is far less existing work at the intersection of model compression and defenses to adversarial attacks. An example of how compression techniques could be adapted for defense against adversarial attacks is given by the defensive quantization technique discussed in [25] and [26]. Future work should consider how the relationships between model compression techniques and robustness to adversarial attacks can be leveraged to create new defensive techniques.